# Title: Beyond Black-Box AI: Interpretable Hybrid Systems for Dementia Care

Matthew JY Kang[1,2], Wenli Yang[3], Monica R Roberts[4], Byeong Ho Kang[3], Charles B Malpas[5,6]

[1] Neuropsychiatry Centre, The Royal Melbourne Hospital, Grattan St VIC 3052, Melbourne, Australia

[2] Department of Psychiatry, The University of Melbourne, Grattan St VIC 3052, Melbourne, Australia

[3] University of Tasmania, Churchill Ave, Hobart, 7005, TAS, Australia

[4] Alfred Mental and Addiction Health, Alfred Health, Commercial Rd, VIC 3005

[5] Department of Medicine, Melbourne Medical School, University of Melbourne, Grattan St VIC 3052, Melbourne, Australia

[6] Melbourne School of Psychological Sciences, University of Melbourne, Grattan St VIC 3052, Melbourne, Australia

**Abstract**

The recent boom of large language models (LLMs) has re-ignited the hope that artificial intelligence (AI) systems could aid medical diagnosis. Yet despite dazzling benchmark scores, LLM assistants have yet to deliver measurable improvements at the bedside. This scoping review aims to highlight the areas where AI is limited to make practical contributions in the clinical setting, specifically in dementia diagnosis and care.

Standalone machine-learning models excel at pattern recognition but seldom provide actionable, interpretable guidance, eroding clinician trust. Adjacent use of LLMs by physicians did not result in better diagnostic accuracy or speed. Key limitations trace to the data-driven paradigm: black-box outputs which lack transparency, vulnerability to hallucinations, and weak causal reasoning. Hybrid approaches that combine statistical learning with expert rule-based knowledge, and involve clinicians throughout the process, help bring back interpretability. They also fit better with existing clinical workflows, as seen in examples like PEIRS and ATHENA-CDS.

Future decision-support should prioritise explanatory coherence by linking predictions to clinically meaningful causes. This can be done through neuro-symbolic or hybrid AI that combines the language ability of LLMs with human causal expertise. AI researchers have addressed this direction, with explainable AI and neuro-symbolic AI being the next logical steps in further advancement in AI. However, they are still based on data-driven knowledge integration instead of human-in-the-loop approaches. Future research should measure success not only by accuracy but by improvements in clinician understanding, workflow fit, and patient outcomes. A better understanding of what helps improve human-computer interactions is greatly needed for AI systems to become part of clinical practice.

# 1. Introduction

Recent years have seen an explosion of artificial intelligence (AI) applications in medicine. Most of these tools follow a dominant trajectory: they leverage machine learning (ML) on big data to produce statistical predictions – risk scores, likelihood percentage (%), 'black-box' biomarkers – that promise early detection or outcome forecasting. While often technically impressive, such AI outputs can be difficult for clinicians and patients to interpret or act upon in a meaningful way. A clinician assessing a patient with memory problems may be presented with a clear probability of a clinical outcome (e.g. "85% risk of conversion to Alzheimer's disease"). Yet without an explanation or guidance, the AI system output for situation means the broader clinical situation is still opaque (other possibilities include depression mimicking dementia, undiagnosed obstructive sleep apnoea, heavy drinking history being withheld). This narrative review examines the limitations of this prevailing paradigm and argues for a hybrid approach that combines statistical learning with expert-curated rules to yield interpretable, clinically actionable outputs.

Large language models (LLMs) have re-ignited the adoption of AI in healthcare. At its core, the promise stems from their ability to produce human-like responses. In the first RCT of physicians using an LLM in real time, Goh et al. found *no significant gain in diagnostic reasoning* when GPT-4 was available compared with conventional resources (Goh et al., 2024). Using an LLM assistant has neither improved diagnostic accuracy nor time efficiency. Although LLM-driven scribing tools have gained popularity, their usefulness as a clinical decision support system has not yet materialised. Attention has now turned to the integration of AI approaches, such as LLM and knowledge-based systems (Table 1; Yang et al., 2025). However, the literature is still missing from the perspective from the clinician, including what is needed for an AI system to be useful in clinical practice.

| Levels of hybrid AI integration | Description | Case Example |
|---|---|---|
| Level 1: Knowledge retrieval & integration | Focuses on retrieving and integrating knowledge sources to improve accuracy and contextualisation whilst reducing hallucination risks. | **Age-adjusted p-tau217 lookup** → The LLM retrieves latest cut-off tables (e.g., >190 pg/mL for ≥65 yrs) and systematic-review snippets on diagnostic accuracy. It adds this evidence to the electronic lab report, so the clinician sees *both* the patient's value (242 pg/mL) *and* the guideline-anchored reference range plus key citations. |
| Level 2: Knowledge utilisation & reasoning | Incorporate reasoning to enhance interpretability, as well as select relevant information to provide context-specific actions | **Hybrid rule + LLM interpretive comment** → Rule engine checks: ↑ p-tau217, ↑ plasma NfL (>2 SD for age), MMSE < 26, no vascular lesions on MRI. It fires a rule: "Probable AD pathology". The |

| | | LLM then drafts a plan: *"Order amyloid-PET or CSF Aβ/tau to confirm; start cholinesterase inhibitor if confirmed; refer to dementia clinic service; counsel on driving & advance care planning."* |
|---|---|---|
| Level 3: Optimisation & specialisation | Self-improvement for long-term optimisation, coupled with the clinician tailoring the system performance for domain-specific needs | **Adaptive brain health monitoring coach** → System ingests each patient's serial investigation reults plus follow-up diagnoses from your memory clinic. Every quarter it fine-tunes its threshold tables and prediction model to local population characteristics (e.g., higher baseline NfL in stroke-referral cohort). Dashboard shows updated PPV/NPV and recommends personalised retest intervals (e.g., 6 months if NfL rising >15 %/yr). Clinicians can override, and their feedback becomes new training data. |

Table 1. Progressive Levels of Hybrid AI Integration into Clinical Workflows. Adapted from Yang et al 2025 (Yang et al., 2025)

This scoping review explores how artificial intelligence is applied in medicine and how it supports the development of new technologies in healthcare, especially in dementia care. The paper aims to give a broad yet focused view of current trends and practical uses of AI in clinical settings. Its major contributions are:

- Providing a structured overview of AI applications across different medical fields.
- Identifying how AI tools enhance emerging healthcare technologies.
- Presenting real-world case examples where AI assists clinicians in context-specific scenarios.
- Examining the main challenges and opportunities in adopting AI to improve patient outcomes.

# 2. Overview of AI tools in medicine

Artificial intelligence (AI) is changing many parts of healthcare. In medicine, AI tools are now used to help with diagnosis, prediction, and treatment planning. These tools can process large amounts of patient data and find patterns that are hard for humans to detect. In dementia care, for example, AI models are being used to predict who might develop cognitive decline, based on scans, speech, genetics, or wearable device data. These advances are promising. But in real clinics, many of these tools are still not widely used. This section examines how medical AI tools are used today, what challenges they face, and how they can be made more useful for doctors and patients.

## 2.1 Limits of Prediction-Only Tools

Contemporary data-driven medical AI tools often emphasise prediction over explanation. Machine learning algorithms can detect patterns from clinical data to predict outcomes like dementia risk. Deep learning models analyse speech (Zolnoori et al., 2023), neuroimaging MRI scans (Nguyen et al., 2023), electroencephalograms (Akras et al., 2025), and wearables (Lown et al., 2020) to classify disease subtypes or forecast progression. These advances are valuable in *research* settings – where they improve diagnostic accuracy and prognostication in controlled studies (Rony et al., 2025). However, the *form* in which these predictions are delivered to clinicians is often a simple risk score or classification label produced by a complex model (McNamara et al., 2024; Nguyen et al., 2023). The output resembles a traditional risk calculator ("this patient has a 30% chance of cognitive decline in 5 years") but without the benefit of a clear rationale.

Another challenge is the lack of consistency across AI models. Different models are trained on different datasets, use different types of input data, and apply different thresholds when producing risk scores or classification labels (Jacobs et al., 2021). For example, one model might rely heavily on MRI imaging, while another uses speech or genetic data. One might flag a patient as a 'high risk' at a 20% threshold, while another uses 40%. This makes it difficult for clinicians to compare results across systems, or to decide which tool to trust. It also raises concerns about fairness and reproducibility, especially when these tools are applied to different patient populations. On top of that, most models are not designed to handle real-world clinical data, which is often incomplete, inconsistent, or recorded differently across settings. Important patient information might be missing, outdated, or entered in free-text notes that the model cannot read. Even small changes in data format can cause errors. These limitations reduce both the accuracy and trustworthiness of prediction-only tools when used outside the lab.

## 2.2 Biomarkers grow fast but guide little

There has also been a wave of novel biomarkers in dementia care, including polygenic risk scores and multi-omic data. Yet studies have thus far either evaluated the performance of a specific biomarker (i.e. phosphorylated tau 271 (p-tau217) above a certain threshold infers a risk of Alzheimer's disease) or used machine learning modelling to identify the best formula that predicts a certain outcome (i.e. combination of polygenic risk score poor cognition infers a high risk of Alzheimer's disease). Despite the promising results of these novel biomarkers, their uptake in clinical practice is lagging.

One key reason is that most of these biomarkers are used to improve statistical prediction but not clinical guidance. They are often added into complex models that increase accuracy, but they rarely change what a clinician does. For instance, knowing that a patient has a high genetic risk score may not be useful unless it comes with a recommended treatment or test. Clinicians may hesitate to act on a number if the meaning is unclear or not linked to approved guidelines. In addition, many of these biomarkers are still being validated. Their performance may differ across age groups, ethnic backgrounds, or comorbid conditions. Without clear standards, clinicians are unsure how to apply them in real-world cases.

## 2.3 Explanation Without Action

To make AI results easier to understand, some researchers build explainable AI (XAI). These models try to show what led to the final decision. They might highlight key data points, like changes in brain structure or pauses in speech. Some use colour maps to mark areas of interest in brain images. Others show how much each input added to the final score. For example, Iqbal et al. (2024) utilised Local Interpretable Model-agnostic Explanations (LIME) and SHapley Additive exPlanations (SHAP) to identify influential linguistic features in language output for Alzheimer's dementia screening, achieving an 80% classification accuracy. Similarly, Altinok (2024) proposed an explainable multimodal fusion approach combining text and speech data using cross-attention mechanisms for dementia detection.

Even with these tools, the problem of interpretation remains. Although techniques like LIME, SHAP, or attention-based models can highlight important features or show how the decision was made, the outputs are often still too technical or abstract for routine clinical use. Clinicians may see which features influenced the model, but they are still left to decide what action to take. This leads to an interpretation gap: they must figure out why the AI model made its prediction and how to integrate that into care. If an algorithm flags a patient as 'high risk' based on subtle patterns in their data, the care provider is left wondering what to do next. (Petch et al., 2022). In other words, the model's job (prediction) is done, but the clinician's job (decision-making) just became harder. Does a 70% likelihood mean we start anti-dementia therapy? Or do we order a confirmatory PET scan? The AI, as it stands, won't tell us. The clinical utility of an AI's output is limited if it does not come with context or actionable guidance. These considerations are especially relevant given that a clinician must take final responsibility for any diagnosis or treatment that emerges from the clinical encounter. If a clinician is unable to understand the process by which the diagnosis or management decision was reached, then they will not be able to defend the decision to the patient, colleagues, or in the context of litigation.

# 3. From Black Boxes to Clinical Judgment

The previous section showed that while AI systems in dementia care are becoming more advanced in predicting outcomes, they often fall short in helping clinicians take the next step. Even explainable models rarely provide the type of reasoning or actionable recommendations that align with how doctors think and make decisions. This creates a persistent interpretation gap, where the model outputs are technically impressive but clinically unclear. To understand why this gap matters, and how to move forward, we must examine the deeper limitations of black-box AI systems, revisit what makes human reasoning essential in medicine, and reflect on past efforts like expert systems that aimed to embed medical knowledge directly into decision support tools. This section explores the trust barriers created by AI models, the strengths of causal and contextual reasoning in human clinicians, and the practical lessons from earlier clinician-centric AI designs.

## 3.1 Black-Box Models: Barriers to Trust and Use

A fundamental criticism of existing medical AI systems is their 'black-box' nature. Complex models often do not provide human-understandable reasoning for their predictions. The internal logic – millions of parameters extracted from training data – is opaque. This lack of interpretability has several negative consequences:

**Undermined Trust:** Clinicians are wary of accepting a life-altering prediction (i.e. diagnosis of dementia) from an algorithm that cannot explain its reasoning (Petch et al., 2022). Moreover, it's hard to know *why* a model made a mistake, making it difficult to correct errors or improve the system. This raises ethical and legal questions around accountability, responsibility, and liability in AI-supported medical decisions.

**Reduced Clinical Insight:** Traditional statistical models, such as logistic regression, provide interpretable coefficients and confidence intervals. For example, the presence of a measurable sign can be associated with the change in odds for a particular clinical outcome. These allow clinicians to learn something about the underlying disease process, for example, how strongly a biomarker is associated with an outcome. Black-box models do not directly offer such insight into disease relationships. They generate results but do not contribute to the clinician's understanding of the patient's condition. Thus, clinicians gain no new understanding of disease process from the AI – they only get a prediction. In fields like dementia where mechanistic understanding is still evolving, this is a missed opportunity.

**False Promise of Explainability:** while some tools use post-hoc explainability methods such as heat maps or feature weights, these explanations may produce plausible sounding but misleading rationales. Ghassemi et al. argue these explanations are often superficial, vulnerable to human confirmation bias, and lack performance guarantees (Ghassemi et al., 2021). Rather than reassuring clinicians, they might increase automation bias – the tendency to trust algorithmic output without clinical reasoning while failing to highlight systemic bias or guide safe care. In large models, LLMs have addressed some weaknesses of traditional machine learning, achieving impressive fluency and adaptability. However, despite their advancements, LLMs do not fundamentally solve the core issues of machine learning. Like other data-driven AI models, LLMs use a similar basic approach to knowledge by learning patterns from large datasets. Their ability to generate text at scale does not inherently grant them reasoning abilities or factual accuracy. They still rely on patterns from massive datasets rather than true understanding. Salvi et al., 2025) note that confident sounding outputs can appear authoritative, making it difficult to detect errors without rigorous quality checks. In summary, the current generation of AI tools, mostly based on predictive analytics and black-box machine learning models, often falls short of delivering final, usable outcomes. They provide *probabilities* instead of *plans*, *flags* instead of *explanations*. This lack of transparency and actionability remains a key barrier to clinical adoption. While many researchers advocate for explainable AI as a pathway to trust and usability, others caution against overstating its current capabilities. To move forward, future systems must go beyond the black box models. They should give not just accurate predictions but also results that clinicians can trust, question, and use in daily practice.

## 3.2 Explanatory coherence and causal reasoning: why humans still lead

Paul Thagard defines explanatory coherence as the degree to which a set of propositions mutually support a causal, consistent account of observed facts (Thagard, 1989). In medicine, clinicians often build clear causal stories by connecting symptoms, test results, and social factors into one testable narrative ("amyloid plaques → synaptic loss → memory decline but consider vascular burden because of long-standing hypertension").

Despite the impressive ability of LLMs to sound and reason like a human, the evidence remains that they still fall short in causal reasoning. (Dettki et al., 2025; Kıcıman et al., 2024; Thagard, 2024). Moreover, the errors made by LLMs are unpredictable and inconsistent (Kıcıman et al., 2024). In a real-world randomised clinical trial using case vignettes, GPT-4 assistance did not improve physicians' diagnostic accuracy or efficiency (Goh et al., 2024), highlighting a gap between fluent text and clinical insight.

The limitations of AI, including LLMs, are particularly problematic in areas of medicine where the exact mechanistic understanding of specific disorders is lacking. This includes Alzheimer's disease, the most common cause of dementia, where experts still rely on theories about the mechanism of pathology (Abubakar et al., 2022; Sheppard & Coleman, 2020). It is unclear how AI can overcome the limitations in these areas, as it will struggle to explain what is currently unexplainable (Rosenbacke et al., 2024). This could lead to attempted explanations that are incorrect. On the other hand, clinicians can acknowledge their limitations in understanding and are not limited by the competing demands of abstraction and granularity levels.

## 3.3 Lessons from Expert Systems: Clinician-Centric AI

It is ironic that decades ago, in the early era of AI, many medical expert systems were explicitly designed to provide *interpretable advice*. For example, MYCIN was developed in the 1970s by AI pioneer Bruce Buchanan to identify bacteria causing infections, recommend antibiotics, and *provide reasoning* (Swartout, 1985). Other early systems – often rule-based – acted as virtual experts, generating explanatory reports or recommendations much like a seasoned clinician would. Another classic example was *PEIRS* (Pathologist's Expert Interpretative Reporting System), a pathologist-maintained rule-based system for interpreting chemical pathology reports (Edwards et al., 1993). It contained ~950 rules and could automatically add commentary to lab results, explaining abnormal patterns and suggesting next steps. Notably, PEIRS was maintained by the domain experts (pathologists) themselves, not by programmers, and has been learning new rules incrementally by the domain experts – a key feature for longevity. The failure and success of rule-based expert systems like MYCIN and PEIRS offer several lessons as we consider the future of AI in dementia care:

**Context is Key:** One of the key strengths of expert systems is their ability to include context directly in their logic. Rule-based systems do not just link variables to outcomes. They also define the conditions that must be met for a conclusion to be valid. For example, a rule might state: *"If the patient is over 65, has elevated CRP, and shows confusion, consider infection-related delirium."* This conditional reasoning mirrors the way clinicians think, balancing multiple variables in specific clinical settings. In dementia care, where differential diagnosis depends heavily on age, comorbid conditions, family history, functional decline, and test results, context is especially important. Without contextual reasoning, even accurate predictions can feel disconnected from clinical reality and are less likely to influence decision-making.

**Ongoing Knowledge Maintenance:** Medicine evolves, and so must AI. The AI system must allow domain experts to directly refine AI's knowledge in a piecemeal fashion, which is a powerful way to keep systems accurate. This process supports adaptability and transparency, allowing clinicians to incorporate new medical findings, guidelines, or diagnostic pathways without retraining the entire system. It also creates an opportunity for continuous learning, where the system reflects not just data trends but clinical judgment. This user-driven maintenance contrasts with one-time ML model training, which may become outdated as new patterns (or biomarkers) emerge. Note that repeated ML model training by an updated dataset is not the solution, as this does not capture the contextual differences between what was previously known and newly learnt. More importantly, it offers no direct mechanism for clinicians to correct or annotate decision logic, nor does it support traceable, case-based updates that are essential for safety, accountability, and trust in clinical environments.

**Interpretable Outputs with Cases:** Rule-based conclusions can be traced back to the conditions that fired them, essentially providing a built-in explanation ("because TSH was high and T4 low, I concluded hypothyroid"). This transparency is similar to a clinician explaining their reasoning and is in stark contrast to black-box nets. This traceability supports clinical trust and lets clinicians challenge, refine, or confirm the logic used. Most deep learning models do not allow this kind of interaction. The success of PEIRS is based on its knowledge update, which is always linked to two counter cases, and it interacts with human experts to avoid the inconsistencies in heuristic human knowledge through cases. This case-based interaction also acts as a safeguard, helping the system resolve ambiguity and refine edge cases over time. It turns each update into a learning opportunity, not just for the system, but for the clinicians involved as well.

**Focused Scope with High Precision:** Many expert systems were narrow in scope (e.g., interpreting lab tests, or advising on a specific disorder) but within that scope, they achieved high diagnostic accuracy and consistency. This suggests that AI need not be grandiose to be useful – even targeted decision support tools can significantly aid clinical practice. In fact, limiting scope may increase reliability by reducing the number of variables the system must handle, which in turn lowers the risk of unexpected behavior or incorrect generalisation. Narrow systems can be rigorously tested and validated against known clinical standards, making them easier to regulate, update, and deploy safely. These systems are also more likely to be accepted by clinicians, who can clearly understand their purpose, limitations, and relevance to specific tasks. In areas like dementia care, where clinical workflows involve multiple stages and data types, such focused tools can support discrete decision points such as initial screening, biomarker interpretation, or treatment planning without attempting to replace the entire diagnostic process.

However, the uptake of knowledge bases, including pure rule bases, has stalled. PEIRS, a rule-based system, has been widely used and has demonstrated the strength of human-in-the-loop approaches to cases. However, its success is limited to a few well-defined, metric-based domains, such as pathology lab system areas. This is due to their brittle and resource-intensive nature. Extracting tacit knowledge from domain experts is labour-intensive and slow. The scope is often narrow as inputs must match pre-determined patterns. Moreover, ongoing maintenance is difficult, especially in the medical space where being up to date with the latest guidelines and best practices is critical for acceptability.

# 4. Hybrid AI: Merging Statistical Learning with Expert Knowledge

## 4.1 Hybrid Intelligence: Strengths-Based Integration of ML and Rules

Given the respective strengths and weaknesses of black-box ML and expert rule-based systems, a compelling path forward is to combine these approaches into hybrid AI systems as shown in Figure 1. In a hybrid model, machine learning algorithms and expert-curated rules are not competitors but collaborators, each contributing where they are strongest.

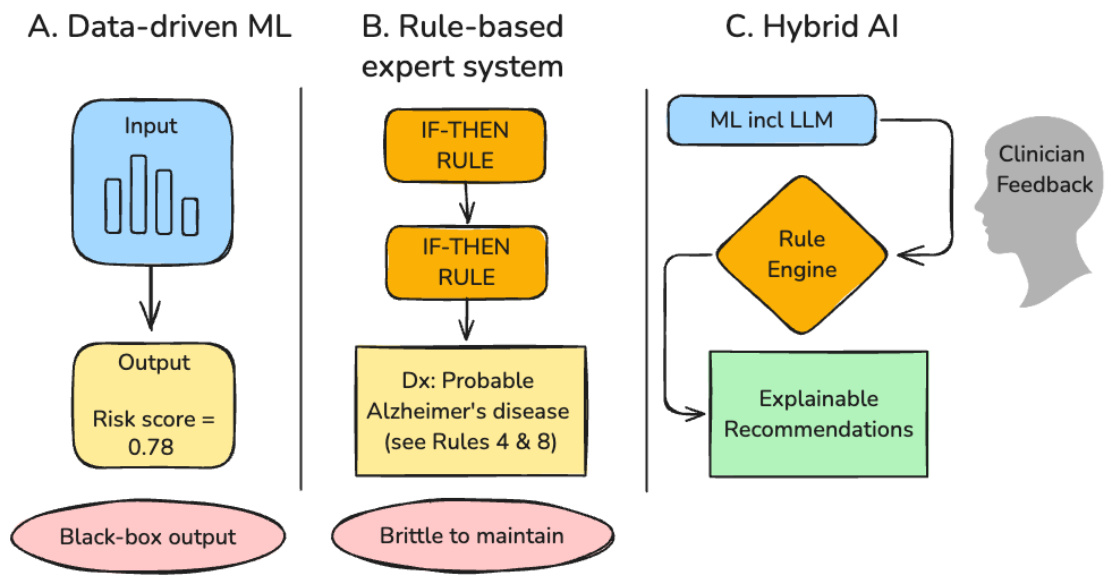

Figure 1. Comparing AI approaches for clinical decision support tool in dementia care

Machine Learning for Pattern Discovery: ML models can discover complex patterns in high-dimensional data that humans might miss (for example, subtle neuroimaging features correlating with early psychosis, or a polygenic risk score that predicts autism). They can continuously learn as more data becomes available, improving predictive accuracy. Given the wealth of new data from novel biomarkers, ML models can help clinicians avoid being overwhelmed by data (Thagard, 1989). These models also enable population-level insights by aggregating trends across thousands of patient records, identifying risks or trajectories not visible at the individual level. In dementia care, for instance, ML systems can surface early combinations of symptoms that correlate with progression to Alzheimer's dementia, helping identify at-risk individuals before clinical symptoms are obvious.

Expert Rules for Domain Knowledge and Contextual Interpretation: Expert systems or rule bases can inject clinical domain knowledge and context directly into the decision process, ensuring that the AI's output aligns with medical reasoning and accepted standards of care. They can encode things like "if the patient's p-tau217 is high *but* they have a history of traumatic brain injury, consider that as a confounding factor rather than immediate Alzheimer's" – a nuance a pure ML model might not capture if that scenario was rare in training data. These systems reflect how clinicians think and reason, often prioritizing causal explanations and guideline-concordant actions. Rules can also incorporate exceptions, thresholds, or red flags that align with real-world judgment (e.g., "Do not consider elevated NfL alone diagnostic if MMSE is normal and patient is under 40"). This ensures that output remains interpretable, auditable, and directly linked to established clinical pathways.

Table 2 summarises the distinct yet complementary roles of machine learning and expert rule-based systems, highlighting how each contributes uniquely to hybrid clinical AI design.

| Component | Machine Learning (ML) | Expert Rule-Based Systems |
|---|---|---|
| Main Strength | Pattern recognition in large, high-dimensional datasets | Embedding clinical knowledge and context-specific rules |
| Data Handling | Learning from structured and unstructured data (e.g. MRI, genomics, speech) | Rely on structured input and predefined rule logic |
| Learning Method | Automatically improves with more data | Updated manually by clinicians or domain experts |
| Typical Output | Risk scores, classifications, predictive labels | Conditional recommendations, explanations, and next-step actions |
| Interpretability | Often opaque (black-box') unless supported by explainability tools | Fully transparent and explainable through rule traceability |
| Use Case Examples | Predicting Alzheimer's risk from neuroimaging and genomics | Flagging exceptions in diagnostic pathways; enforcing guideline-based care |
| Limitations | May miss context or rare scenarios; hard to explain decisions | Brittle in face of novel data; difficult to scale across domains |
| Best Applied When | Handling complex, high-volume data requires statistical insight | Needing clear clinical logic, guideline compliance, or clinician auditability |

Table 2. Complementary Strengths of Machine Learning and Expert Rule-Based Systems in Clinical AI

## 4.2 Example Workflow: Hybrid AI in Dementia Care

A hybrid AI might work as follows in a dementia context: An ML model analyses a patient's data (symptoms, cognitive profile, neuroimaging, fluid biomarkers) and produces an initial assessment say, a probability that the patient's cognitive profile fits Alzheimer's disease. This probability and the pertinent features (perhaps the model's top predictors) are then fed into an expert rule-based engine. The rule engine contains medical knowledge such as diagnostic criteria, differential diagnoses, and management guidelines. Based on the output of the model output *and* all available clinical data, the rule engine generates an *interpretable report*. (Table 3)

| Interpretation | *The patient's profile suggests a likelihood of Alzheimer's disease. Key factors: elevated p-tau217 and APOE4 genotype.* |
|---|---|
| Explanation | *This biomarker profile is highly suggestive of Alzheimer pathology in the context of memory impairment.* |

|  | *Differential diagnoses (e.g. chronic traumatic encephalopathy) should be considered given the patient's history of prior head injuries, although the biomarker pattern here is more specific to AD.")* |
| --- | --- |
| Suggested plan | *Confirm with amyloid PET imaging; if positive, begin cholinesterase inhibitor therapy and refer to Alzheimer's clinical trials.* |

Table 3. Example of a contextualised hybrid AI model response for clinicians

Such output provides *the risk prediction in context*, along with next steps and caveats, much like an experienced specialist's consultation note. The statistical model contributes the raw predictive power and pattern recognition, while the expert system ensures the output makes clinical sense and is actionable (suggesting confirmatory tests or treatments rather than leaving the physician with just a number).

## 4.3 Implementing and Embedding Hybrid AI into Clinical Practice

Technically, implementing hybrid systems is becoming increasingly feasible. Knowledge base systems can act as a 'wrapper' around machine learning models – essentially using an expert rule layer to catch and correct the ML model's errors or adjust its outputs. Another approach is to have ML algorithms suggest new rules – for example, by identifying interesting feature combinations – which human experts can then validate and incorporate into the rule base (thus continuously enriching the knowledge base with data-driven insights). These data-driven explanations, while not perfect, can be used as intermediate inputs for expert review. One could imagine an iterative loop: the ML model flags a pattern, the expert system explains it in a preliminary way, a clinician user corrects or refines that explanation, and this feedback is fed back into improving both the model and the rule base. In effect, the system would learn *and* accumulate knowledge, side by side. If successful, the payoff would be significant: AI tools that not only predict but also **p**rescribe or explain, thereby truly assisting in clinical decision-making rather than merely adding another datapoint to ponder.

Alongside technical design, the successful integration of hybrid AI depends on its acceptability within clinical practice. The clinician's resistance against data-driven algorithms is not new. In 1954, Paul Meehl's work 'Clinical versus Statistical Prediction' caused significant controversy (Meehl, 1954). Reviewing several head-to-head studies, Meehl showed that actuarial formulas equaled or surpassed clinicians in the majority of cases. Although some first saw Meehl's work as an "attack upon the clinician," he later explained that data-driven actuarial methods can actually help. They can save the clinician's time, allowing them to focus on tasks where human judgement is essential. These include rare or unusual features, open-ended problems, and reasoning guided by theory (Meehl, 1967). These early findings, the limitations of the feature-based data-driven approach, were not well recognised during the growth of AI. The improvements in data-driven methods and computing power will not overcome the fundamental issues in this approach.

Subsequent work has clarified *how* to keep the clinician in the loop without losing actuarial accuracy. Westen and Weinberger demonstrated that once a clinician's tacit impressions are *structured and quantified*, it can improve diagnostic accuracy (Westen & Weinberger, 2005). Moreover, clinicians are better adapted to provide explanatory coherence to any rules or logics learnt from machine learning. This enables ML-derived data to be converted to a digestible form to clinicians and patients. The experienced clinician considers the broad and unique context – seeing the forest for the trees and identifying "broken leg" conditions under which the results of algorithmic predictions should be put to one side (Meehl, 1967).

Paralleling this is 'human-in-the-loop' computing (Budd et al., 2021), where clinicians can be presented with data-driven learnings that is made transparent by LLM, which is then used to define further rules and cornerstone cases for the Hybrid AI model (Figure 2) to continue its active learning. In effect, structured judgement converts 'intuition' into labelled data that a learning system can ingest. The result is a division of labour Meehl could endorse: the LLM supplies scalable statistical aggregation, the rule base captures mid-level clinical prototypes, and the clinician provides the theory-guided exceptions that keep the system both accurate *and* meaningful.

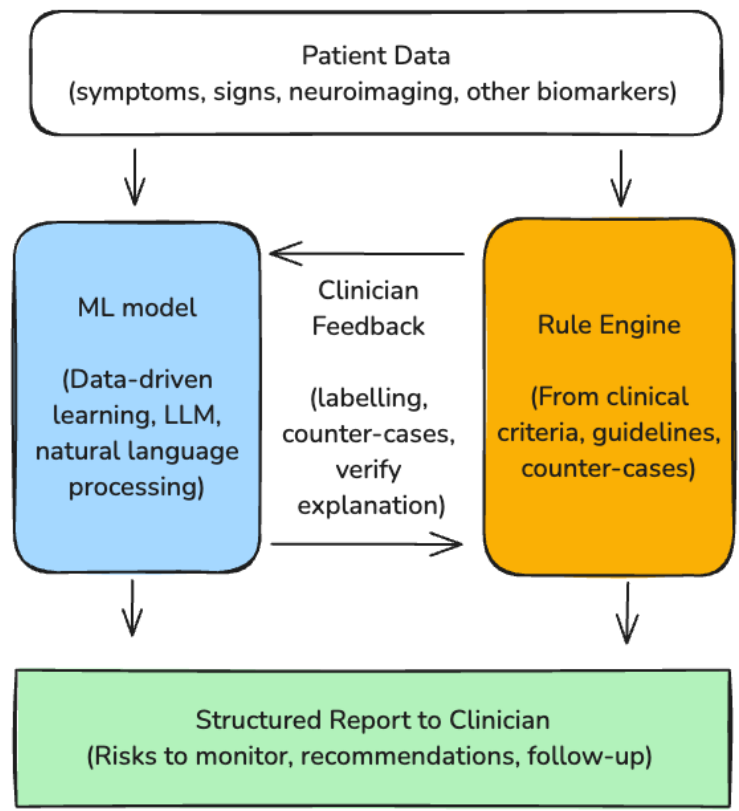

Figure 2. Hybrid AI workflow for dementia care with human-in-the-loop feedback

## 4.4 Implementation as Digital Therapeutics

Digital therapeutics (DTx) are software-based interventions that prevent, treat or manage disease under clinical-grade quality and regulatory oversight are beginning to fill the 'actionability gap' between AI predictions and day-to-day care. Their scalable and flexible nature is particularly appealing in a resource-scarce healthcare system. There is an urgent need for new technology to build the existing capacity of clinicians, given the increasing rates of dementia worldwide which is outpacing the growth of the healthcare workforce (GBD 2019 Mental Disorders Collaborators, 2022; Livingston et al., 2020). DTx can allow clinicians to focus on complex tasks such as diagnosing, formulating, and personalising their care.

Consider, for example, a person presents with memory impairment and functional decline. Despite gold standard assessment by the clinician with the aid of a hybrid AI model, it is still uncertain whether this person will develop dementia at this early stage. During the assessment the hybrid AI model can identify their relevant modifiable risk factors for dementia, such as hypertension, high cholesterol, and smoking, as defined by the 2024 Lancet Commission (Livingston et al., 2024). These DTx tools, including lifestyle intervention-based DTx (Tang et al., 2025), can then recommend specific interventions for each risk factor (i.e., motivational interviewing for smoking cessation, education and monitoring for diet and cholesterol control).

The integration between hybrid AI and DTx tools creates a continuous loop. The hybrid AI system can ingest ongoing input from DTx tools such as progress data, adherence, or biomarker feedback and use this real-time information to re-calculate dementia risk and refine the individual's prognosis. This dynamic feedback loop allows clinicians to monitor how lifestyle changes are influencing risk and adapt care accordingly. It also promotes structured, evidence-based decisions, helping to reduce variability and bias in clinical judgment. In this way, DTx becomes a critical bridge between predictive insight and actionable care. The table below summarises how DTx fits into a hybrid AI-supported dementia care workflow and highlights each contributing at different stages to support proactive, adaptive, and personalised dementia care.

| Stage of Care | Hybrid AI Function | DTx Role | Clinical Benefit |
| --- | --- | --- | --- |
| Initial Risk Assessment | Identify risk probability based on symptoms, biomarkers, and clinical data | Not yet involved at this stage | Establishes baseline risk |
| Risk Factor Identification | Flag modifiable factors (e.g. smoking, hypertension, cholesterol) | Select appropriate intervention pathways (e.g. diet, physical activity modules) | Focused risk reduction targeting patient-specific needs |
| Intervention Planning | Recommend structured follow-up and monitoring protocols | Deliver personalised behaviour-change interventions through apps or digital coaching | Scalable delivery of low-intensity care |
| Dynamic Monitoring & Feedback | Update predictions based on patient-reported outcomes or device data | Report adherence, biometric trends, or side effects back to the AI system | Enables real-time adaptation of care plans |
| Outcome Evaluation and Refinement | Adjust prognosis and next steps based on updated information from DTx systems | Provide feedback loop for continuous learning and optimisation of interventions | Informs clinical decision-making and supports long-term tracking |

Table 2. Role of Digital Therapeutics in Hybrid AI-Supported Dementia Care

# 5. Challenges and Design Strategies for Hybrid AI

Hybrid AI systems offer significant potential, but their success in clinical settings depends on how well they are implemented and adopted. This section outlines both technical and cultural challenges, followed by forward-looking strategies and evaluation approaches for sustainable integration.

## 5.1 Design and Implementation Challenges

Hybrid AI systems promise to combine the best of machine learning and expert reasoning, offering both predictive power and clinical transparency. However, transitioning from predominantly black-box, predictive AI to hybrid, interpretable systems will pose challenges. These challenges span system design, human interaction, knowledge maintenance, and clinical safety. Each must be carefully addressed to ensure hybrid AI becomes a trustworthy and usable tool in medical decision-making.

One major challenge is system integration complexity. Integrating rule-based reasoning with data-driven learning can be complex, especially when their outputs conflict, it becomes hard to reconcile the reasoning. There is a risk of information overload – clinicians will not want an AI that produces a two-page verbose explanation for every case either. Striking the right balance between brevity and completeness in AI-generated reports will be important. Techniques from human-factors engineering and cognitive science (e.g. how doctors read and use consultation notes) should inform how AI outputs are designed.

Another challenge is knowledge maintenance. No matter how easy the system allows clinicians to maintain rule bases with relative ease, it still requires time and effort, which are often in short supply. A solution could be to leverage existing clinical guidelines as a starting knowledge base for the expert system component. As guidelines update (say new criteria for mild cognitive impairment, or new treatment pathways for depression), those changes should propagate into the AI's rule layer. There will need to be governance to ensure the knowledge base remains evidence-based and free of biases. One advantage of a hybrid approach is that any biases or errors in the rule-based component are easier to audit and correct than those buried in a black-box model. The transparent nature of rules – each can be inspected and linked to a justification – aligns with the need for accountability in clinical practice.

The third challenge is bias and clinical safety. Both ML and rule-based systems are vulnerable to embedded biases, whether from skewed training data or oversimplified rules. In hybrid AI, where decisions may affect patient care, transparency is essential. Every recommendation must be open to inspection, and clinicians should be able to trace the reasoning behind it. This supports accountability and ensures that the AI system adheres to safe and accepted clinical standards.

## 5.2 Adoption, Collaboration, and Evaluation

Adopting hybrid AI systems requires a cultural shift in clinic practice. Clinicians must be willing to engage with AI as a collaborator. Instead of seeing the AI as a black-box that outputs a number, clinicians would interact with the AI's explanation, potentially give feedback, and incorporate its suggestions into care plans. This might feel more natural to clinicians, as the AI would be speaking their language (clinical reasoning). Training and experience will be needed to ensure clinicians understand the capabilities and limits of these systems.

Multidisciplinary collaboration is critical. The development of hybrid AI for healthcare should involve data scientists, domain experts, and knowledge engineers. In many cases, clinicians themselves may need to take on the role of 'knowledge engineers', especially under frameworks like ripple down rules (RDR). Such collaboration echoes the collaborative approach advocated in LabWizard's implementation – pathologists working closely with clinical colleagues to capture specialist knowledge in rules, thereby improving decision support.

The incorporation of natural language processing into hybrid frameworks is gaining increasing attention. One can imagine an LLM (trained in medical texts) being used to generate the first draft of an explanation or management plan, which is then vetted and modified by a rule-based system or a human expert. For example, the ML model predicts a high relapse risk for a psychotic patient; an LLM could draft a paragraph explaining possible reasons and suggesting a treatment intensification, and the expert system layer ensures factual accuracy and alignment with guidelines (e.g., checking that the suggested medications are appropriate given the patient's profile). This kind of neuro-symbolic AI (Colelough & Regli, 2025; Garcez & Lamb, 2023) is an active area of research and could accelerate the move toward AI that is not just smart, but able to articulate its reasoning.

Finally, evaluation of AI systems should broaden beyond traditional metrics like accuracy or area-under-curve for predictions. We should assess clinical utility: Does AI's output help clinicians make better decisions or improve patient outcomes? For hybrid systems, this might involve simulation studies, where clinicians are given AI-generated reports in complex cases and their decisions (or diagnostic accuracy) are measured. Early studies in the lab medicine domain have already shown improvements in decision-making with interpretative reporting. Similar evaluations in dementia care could demonstrate the value of interpretable AI assistance. In the long run, the success of medical AI will be measured not by how complex the algorithms are, but by *how much real-world impact they have on patient care*. Interpretability and actionability are key to bridging that last mile to clinical adoption.

## 5.3 Levels of Hybrid AI Integration into Clinical Workflow

Hybrid AI systems can be introduced into clinical practice at varying degrees of sophistication. These levels reflect gradual progression from simple augmentation to complex adaptive support:

- **Level 1: Knowledge Retrieval & Integration:** At this foundational level, hybrid AI systems focus on retrieving and aligning relevant external knowledge with patient-specific data. The goal is to improve contextual accuracy while reducing risks like misinformation or hallucination. By embedding evidence-based references such as biomarker thresholds or diagnostic criteria directly into reports, the system enhances clinical interpretability without altering the core data. This supports clinicians by adding immediate, relevant context to lab values and predictive scores.
- **Level 2: Contextualised Reasoning & Decision Support:** This level adds structured reasoning to the system's capabilities. The AI not only identifies key patterns but also interprets them using encoded clinical logic. Rules are triggered based on combined features across modalities such as cognitive scores, imaging, and biomarkers. These trigger condition-specific insights which are translated into suggested actions or care plans using natural language generation. The result is a preliminary interpretable summary that supports clinical judgment while remaining grounded in structured logic.
- **Level 3: Adaptive Optimisation & Specialisation:** At this advanced level, hybrid AI systems adapt over time to local data, patient populations, and clinical workflows. They incorporate continuous feedback from clinicians, updates from patient outcomes, and shifts in population health trends. This allows the system to personalize thresholds, refine its predictive accuracy, and recommend tailored follow-up actions. Clinician input becomes part of the feedback loop, guiding model adjustments and knowledge base updates to ensure relevance, safety, and effectiveness in real-world practice.

# 6. Future Directions and Research Opportunities

The integration of artificial intelligence into clinical practice is still in its formative stages. This review has outlined the current limitations of prediction-only models and highlighted the promise of hybrid systems that combine machine learning with expert-derived knowledge bases, further enhanced by LLMs. However, hybrid AI tools will only be as good as the ecosystem around it. Therefore, several key directions remain open for further research and development:

## 6.1 Clinician-Centric Interfaces to Maintain Knowledge Bases

Many AI tools fail not because of poor model performance but because they do not fit well into clinical workflows. Future work should focus on building user interfaces that give clear, context-based explanations, let clinicians make changes, and support two-way learning so the AI can adjust based on clinician feedback over time. Ideally, the tool interface will learn which explanation formats each clinician prefers (heat-map, short text, citation list) and default to that view. Collaboration with user interface designers and frontline clinicians is essential to ensure that interfaces enhance rather than impede decision-making.

To keep expert rule systems clinically relevant, we need scalable methods to extract, validate, and maintain medical knowledge bases. Given that active-learning frameworks have already demonstrated success (Lawley et al., 2024), a similar approach can be paired with LLMs for rule curation to allow clinicians to further refine the knowledge base. Further automation can be achieved by incorporating updated guidelines or trial findings.

## 6.2 Making AI Honest about its Blind-Spots

Clinicians also need to know when the algorithm itself is likely to slip for their specific patient scenario. These are what Meehl referred to as the edge cases, empty cells, and situations where the prediction is at higher risk of being wrong. AI tools for clinical care must be able to alert clinicians to these vulnerable situations so that human expertise can step in early. This is in line with the FDA's Guiding Principles for Good Machine Learning Practice (U.S. Food and Drug Administration et al., 2021).

López et al. (2025) recently reviewed the different methods of "uncertainty quantification" of machine learning applications in healthcare. Notably, the authors highlight the lack of clinician input in the existing application of quantifying uncertainty of AI models, especially in the deployment of AI tools, which is crucial for clinical translation. The findings underscore the need for collaboration between clinicians and AI researchers.

This is where the use of cornerstone cases can help clinicians identify why the AI system is more uncertain about their specific case. By comparing the system-flagged "uncertain" case with the cornerstone case it is basing its judgement off, clinicians can quickly identify the "empty cell" or "broken leg" factor that is causing the uncertainty. Furthermore, the new uncertain case can then be recycled and further refine the knowledge base.

However, caution must be used in how this uncertainty is flagged to clinicians, given the risk of alarm fatigue and inducing undue doubt. The clinician must also be cautious in how this uncertainty is communicated to patients and families, to avoid causing more anxiety during a patient-facing consult that is already emotionally charged.

## 6.3 Evaluating Clinical Impact and Barriers in Real-World Settings

Blood-based biomarkers such as p-tau217 are becoming more accessible in dementia care. Yet their integration into practice is limited by interpretability and variability. Future AI systems must not only detect abnormal patterns but also contextualise them. Hybrid AI tools that combine biomarker trajectories with clinical context and decision rules offer a way to operationalise these emerging data sources in routine care. Rules can encode confounders (e.g., chronic kidney injury), while an LLM derives probabilistic progression curves. This dual layer converts a "high" lab value into a patient-specific plan, such as confirmatory amyloid PET vs. watchful waiting.

RCTs of AI tools remain rare, and even fewer are conducted in settings that reflect the complexity of real-world care. Future research should prioritise pragmatic trials (Jin et al., 2024) that assess not just predictive accuracy, but also evaluate 1) clinician trust and adoption, 2) changes in changes in diagnostic or care decision, 3) patient outcomes, and 4) unintended harms including automation bias. Such studies should also include diverse populations, particularly underrepresented groups who may face higher risks of misclassification by AI trained on majority-biased datasets.

Despite promising results of AI tools, it has been scantly adopted in clinical settings (Scott et al., 2024). This utilisation gap suggests that accuracy metrics alone do not translate into bedside adoption. There is a need for further implementation science investigating the cognitive and psychological barriers that hybrid AI must overcome. A key solution will be the co-design of AI systems by developers and clinicians together in adaptive developmental cycles to improve trust, usability and integration into clinical practice.

## 7. Conclusion

This review has summarised the current state of AI tools for dementia care and demonstrated that pure prediction is not enough. Clinicians require transparent reasoning, context-specific recommendations and an indication of uncertainty to be able to adopt these tools into day-to-day care. Historical lessons from expert systems and actuarial prediction highlight the enduring value of human judgement. In order for AI tools to be accepted into clinical care, the knowledge exchange must be bi-directional between the clinician and the AI system.

We have highlighted the progressive levels of hybrid AI integration, mapped DTx workflows and outlined a detailed research agenda. From clinician-centric AI tools to uncertainty signalling, this paper offers a practical roadmap towards AI tools that can translate into clinically actionable aids. Future progress will rely less on small improvements in prediction accuracy and more on strong collaboration. This collaboration should embed AI into the clinician-patient relationship, helping improve diagnostic accuracy, personalise interventions and enhance outcomes in dementia care.